\documentclass[a4paper,english]{rnti}

\usepackage[T1]{fontenc}
\usepackage[utf8]{inputenc}
 \usepackage{amsmath}
\usepackage{graphicx}
\usepackage{algorithm}
\usepackage{algorithmic}
\usepackage{amssymb}
\usepackage{caption}
\usepackage{subcaption}
\usepackage{fvextra}

\titrecourt{DACE For Railway Acronym Disambiguation}
\titre{DACE For Railway Acronym Disambiguation}

\nomcourt{E. Hribach et al.}

\auteur{
El Mokhtar Hribach\affil{1}\textsuperscript{\textdagger},
Oussama Mechhour\affil{2}\textsuperscript{\textdagger}\\
Mohammed Elmonstaser\affil{3}\textsuperscript{\textdagger},
Yassine El Boudouri\affil{4}\textsuperscript{\textdagger},
Othmane Kabal\affil{1}\textsuperscript{\textdagger}
}

\affiliation{
\affil{1}
Nantes University, LS2N, Nantes 44300, France\\
el-mokhtar.hribach@univ-nantes.fr, othmane.kabal@univ-nantes.fr\\

\affil{2}
CIRAD, F-34398 Montpellier, France\\
oussama.mechhour@cirad.fr\\

\affil{3}
Groupe SII, France\\
mohammed.elmontaser@sii.fr\\

\affil{4}
Univ. Lille, CNRS, Centrale Lille, UMR 9189 CRIStAL, F-59000 Lille, France\\
yassine.el-boudouri@univ-lille.fr
}

\resume{
Acronym Disambiguation (AD) is a fundamental challenge in technical text processing, particularly in specialized sectors where high ambiguity complicates automated analysis. This paper addresses AD within the context of the TextMine’26 competition on French railway documentation. We present DACE (\textbf{D}ynamic Prompting, Retrieval \textbf{A}ugmented Generation, \textbf{C}ontextual Selection, and \textbf{E}nsemble Aggregation), a framework that enhances Large Language Models through adaptive in-context learning and external domain knowledge injection. By dynamically tailoring prompts to acronym ambiguity and aggregating ensemble predictions, DACE mitigates hallucination and effectively handles low-resource scenarios. Our approach secured the top rank in the competition with an F1 score of 0.9069.
}

\summary{
La désambiguïsation d'acronymes est un défi fondamental dans le traitement de textes techniques, en particulier dans les secteurs spécialisés où une forte ambiguïté complique l'analyse automatique. Cet article aborde ce problème dans le contexte de la compétition TextMine’26 portant sur la documentation ferroviaire française. Nous présentons DACE (\textbf{D}ynamic Prompting, Retrieval \textbf{A}ugmented Generation, \textbf{C}ontextual Selection, and \textbf{E}nsemble Aggregation), un cadre méthodologique qui améliore les grands modèles de langage (LLM) grâce à un apprentissage en contexte adaptatif et l'injection de connaissances externes. En ajustant dynamiquement les prompts selon l'ambiguïté des acronymes et en agrégeant les prédictions d'ensemble, DACE atténue les hallucinations et gère efficacement les scénarios à faibles ressources. Notre approche a obtenu la première place de la compétition avec un score F1 de 0,9069.
}

\begin{document}

\renewcommand{\thefootnote}{\fnsymbol{footnote}}
\footnotetext[2]{All authors contributed equally to this work.}
\renewcommand{\thefootnote}{\arabic{footnote}}

\section{Introduction}
This work was conducted within the context of the TextMine'26 competition \citep{defi-text-mine-egc-2026}, organized by the \textit{Extraction et Gestion des Connaissances} (EGC) association. The competition aims to confront the scientific state of the art with text-mining challenges faced by industry. This year's edition was proposed by SNCF (French National Railway Company), which is interested in the automatic processing of acronyms.

This specific industrial focus highlights the broader necessity of Acronym Disambiguation (AD), a fundamental problem in natural language understanding and information extraction. Acronyms are ubiquitous in technical and industrial domains, where they provide compact references to processes, locations, organizations, roles, equipment, software components, and more. However, the very concision that makes acronyms useful also creates significant ambiguity: many acronyms are highly polysemous \citep{zahariev2004acronyms}, and their intended meaning depends tightly on domain, subdomain, and local textual context. In safety-critical and highly regulated sectors such as railways, resolving this ambiguity is essential for clear communication, reliable knowledge sharing, and downstream automation (e.g., maintenance planning, compliance checking, and technical search) \citep{kong2024use}.

Historically, approaches to AD have ranged from simple pattern-based and rule-driven heuristics \citep{schwartz2002simple,park2001hybrid} to supervised classification using handcrafted features \citep{okazaki2006building}, and more recently to representation-based methods using contextual embeddings \citep{li2015acronym} and fine-tuned transformers \citep{veyseh2020acronym}. While supervised models can achieve strong performance where abundant, high-quality annotations exist, they struggle to generalize in low-resource settings or to adapt to new domains. Large language models (LLMs) offer a complementary paradigm: through prompt engineering and in-context learning, LLMs can perform complex linguistic reasoning and disambiguation with few or even zero labeled examples \citep{kugic2024disambiguation}. Their flexibility opens new opportunities for domain adaptation, but also raises questions about how to design prompts, select informative demonstrations, and incorporate structured domain knowledge to maximize reliability.

In this paper, we present \textbf{DACE} (\textbf{D}ynamic Prompting, Retrieval \textbf{A}ugmented Generation, \textbf{C}ontextual Selection, and \textbf{E}nsemble Aggregation), a practical framework for acronym disambiguation in railway technical texts that utilizes the strengths of modern LLMs while addressing the specific challenges of the TextMine'26 setting. The system is architected around four integrated components:

\begin{enumerate}
    \item \textbf{Dynamic Prompting} that adjusts the LLM prompt's complexity based on the ambiguity or difficulty of each acronym instance.
    \item \textbf{Retrieval Augmented Generation (RAG)} to bring external authoritative railway glossaries and ontology fragments into the prompting process, helping the model ground ambiguous candidates.
    \item \textbf{Contextual Selection} for choosing a small set of informative, class-balanced in-context examples adapted to the input text.
    \item \textbf{Ensemble Aggregation} (Ensemble Learning) scheme that combines outputs from different model configurations to improve stability and reliability.
\end{enumerate}

Together, these components are designed to combine low-data efficiency, domain specificity, and operational robustness.

We evaluate our approach on the TextMine’26 corpus and analyze its behavior across frequent, rare and highly polysemous acronyms. The results show that dynamic prompting and careful contextual selection substantially improve disambiguation accuracy in low-resource regimes, and that incorporating external domain knowledge and multi-model aggregation yields further gains in consistency and error resilience.

\section{Related Work}

\subsection{Acronym Disambiguation}
Acronym disambiguation has been studied extensively in natural language understanding since early heuristic approaches relied on pattern matching and parenthetical extraction rules. Classic work by \cite{schwartz2002simple} formalized dictionary-based identification of long forms from local contextual information, providing a rule-driven baseline that required no supervised data.

Subsequent research framed acronym resolution as a supervised classification task or as an instance of word-sense disambiguation \citep{chen2023gladis}. Early systems applied SVMs, Naive Bayes classifiers, and handcrafted linguistic features to match acronyms to candidate expansions. These approaches relied on curated acronym dictionaries, which were typically built from annotated corpora or rule-based extraction pipelines. The availability of domain-specific corpora, for example biomedical and scientific text collections, enabled more robust evaluation settings \citep{chen2023gladis}. With the introduction of word embeddings, context and candidate long forms could be represented in a shared semantic space, enabling models to rely on distributional semantics rather than simple token overlap. This capability was first demonstrated in \cite{wu2015clinical} and later refined by embedding-based classifiers and similarity-scoring approaches \citep{song2022t5}.

The advent of transformer-based language models further accelerated progress in acronym disambiguation. Following the release of the SciAD benchmark in 2020, most notably SciBERT-variants consistently set new performance baselines on scientific acronym datasets \citep{pan2021bert, zhong2021leveraging, chen2023gladis}. Research also began to explore cross-domain acronym expansion, aiming to generalize beyond tightly curated disciplinary corpora. MadDog \citep{veyseh2021maddog}, for example, integrates acronyms drawn from heterogeneous domains to support more robust disambiguation, while the large-scale GLADIS resource \citep{chen2023gladis} broadened the empirical foundation of the field by providing over one million annotated acronym instances.

Despite this, domain shift and low resource settings remain notable challenges. Supervised transformers perform strongly when training and test distributions are aligned, yet their accuracy degrades in specialized or underrepresented domains. Unsupervised and weakly supervised approaches, which use unlabeled occurrences or document level semantic consistency, attempt to reduce this dependency but often fall short of fully supervised methods \citep{song2022t5}. These limitations motivate the exploration of more flexible strategies that can adapt to new technical domains with minimal annotation effort.

Recent work therefore investigates LLMs as zero-shot or few-shot acronym resolvers. \cite{kugic2024disambiguation} showed that Gpt-4 \citep{achiam2023gpt} achieves competitive performance on the English CASI benchmark without fine-tuning, indicating that substantial acronym knowledge is encoded during pretraining. The same study reported significant degradation on German and Portuguese acronyms, which illustrates sensitivity to domain and language variation. Smaller open source models such as LLaMA 2 \citep{touvron2023llama} exhibited even lower stability. Few-shot strategies that generate synthetic examples for downstream embedding based classifiers, as explored in \cite{kugic2025embedding}, achieved moderate performance but did not fully resolve ambiguity. These results suggest that LLMs provide strong generalization capabilities, although reliable adaptation to highly specialized jargon may require additional mechanisms.

\subsection{In-Context Learning and Prompt Engineering} 
LLMs have established in-context learning as a flexible alternative to task-specific fine-tuning. In this paradigm, the model conditions its generation on instructions or a limited set of examples provided within the prompt context, performing the task without parameter updates. Prompt engineering has therefore become an active research area, since phrasing, structure, and example selection can substantially affect output quality \citep{sumanathilaka2025prompt}. For tasks such as Word Sense Disambiguation (WSD), LLMs exhibit strong latent linguistic knowledge. \cite{sainz2023language} showed that WSD can be reformulated as a textual-entailment problem, allowing LLMs to judge whether a sense fits a given context using only natural-language prompts.

While zero-shot prompting can establish a viable baseline, it often suffers from high variance and hallucination in specialized domains. In contrast, performance and consistency typically improve significantly with few-shot prompting. This improvement arises because in-context examples serve a dual purpose: they not only provide ground-truth knowledge but also constrain the output space and demonstrate the desired reasoning format \citep{min2022rethinking}. Furthermore, recent work indicates that the selection of these examples is critical; rather than using random demonstrations, retrieving examples that are semantically similar to the target input maximizes the model's ability to generalize to new instances \citep{liu2022makes, rubin2022learning}. This principle applies directly to acronym disambiguation. If an acronym has a dominant expansion, unbalanced prompts encourage the model to default to that expansion. In contrast, prompts that intentionally include rare senses improve context driven reasoning and reduce frequency bias.

Retrieval augmented prompting offers a complementary strategy by injecting external knowledge directly into the model's inference context. In highly specialized technical domains, standard LLMs often fail to recall long-tail entities, leading them to misinterpret domain-specific acronyms or hallucinate plausible but incorrect expansions \citep{kandpal2023large, ji2023survey}. Retrieval mechanisms mitigate this by introducing authoritative definitions and usage examples from verified glossaries. The RAG framework has proven highly effective for such knowledge-intensive tasks \citep{lewis2020retrieval} and is particularly well-suited for acronym disambiguation. By explicitly providing candidate expansions and their definitions, retrieval not only constrains the generation to valid terms but also enhances the verifiability and interpretability of the model's reasoning \citep{shuster2021retrieval}.

\section{Proposed Approach}

\subsection{Problem Formulation}
Formally, we define the acronym disambiguation task as follows. Let $a_i$ denote a specific acronym occurrence within a surrounding contextual text $t_i$. Associated with this acronym is a candidate set $O_i = \{ O_{i_1}, \ldots, O_{i_k} \}$ containing $k$ potential long forms. The objective is to predict the ground-truth label $y_i$, which identifies the correct expansion(s) from $O_i$. This label set may contain a single correct option, multiple valid options, or be empty if no candidate matches.

We denote the training set of $\nu$ annotated examples as $E_{\text{train}} = \{ ((a_i, t_i, O_i), y_i) \}_{i=1}^\nu$. Conversely, the test set consists of $\mu$ instances $E_{\text{test}} = \{ (a_j, t_j, O_j) \}_{j=1}^\mu$, where the labels are unknown. The test set is stratified into two categories: acronyms seen during training ($C_{\text{test} \cap \text{train}}$) and unseen acronyms unique to the test set ($C_{\text{test} \setminus \text{train}}$).

Ultimately, the acronym disambiguation system is modeled as a mapping function $f_{\text{AD}}$ that predicts the label for a given test instance:
\[
f_{\text{AD}} : (a_j, t_j, O_j) \mapsto \hat{y}_j.
\]

\subsection{The DACE Framework Overview}
\begin{figure}[ht]
    \centering
    \includegraphics[width=\textwidth]{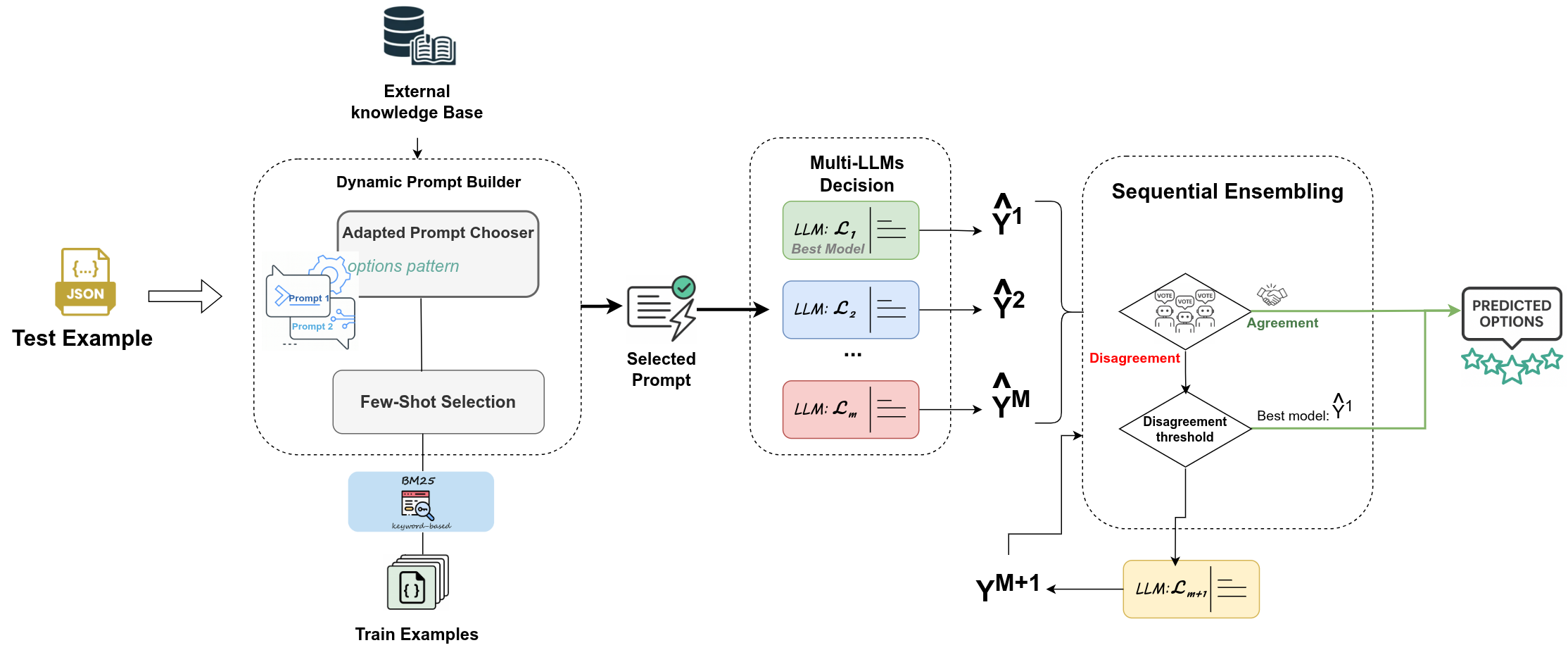}
    \caption{Architecture of the DACE framework.}
    \label{fig:dare_archi}
\end{figure}

Building on the formal task definition, we introduce \textbf{DACE}, a modular framework designed to enhance LLMs performance through structured in-context learning. The architecture orchestrates four synergistic components: (1) \textbf{Dynamic Prompting} to adapt instruction complexity; (2) \textbf{Retrieval Augmented Generation} to inject domain knowledge; (3) \textbf{Contextual Selection} to provide relevant demonstrations; and (4) \textbf{Ensemble Aggregation} to stabilize predictions. An overview of the full pipeline is presented in Figure \ref{fig:dare_archi}. The following subsections detail the implementation of these modules.

\subsection{Dynamic Prompting}
\label{sec:dynamic_prompting}
To accommodate the structural heterogeneity of acronym usages in the TextMine'26 dataset, our system abandons static instruction templates in favor of a dynamic prompt builder. This module constructs the final inference context by synthesizing the retrieved domain knowledge (Section \ref{sec:retrieval}) and the selected few-shot demonstrations (Section \ref{sec:selection}). Crucially, it employs a switching mechanism that alters the instruction strategy based on the specific difficulty profile of the input instance.

Preliminary error analysis indicated that a ``one-size-fits-all'' instruction fails to generalize across both common and rare acronyms, particularly when candidate options exhibit high lexical overlap. To mitigate this, we define a switching function that selects between $p=2$ template configurations based on the acronym's training status and candidate similarity.

\begin{itemize}
    \item \textbf{Template A (Standard Context):} This is the default configuration, applied when the acronym is present in the training set ($a_j \in C_{\text{test} \cap \text{train}}$). In these cases, the in-context examples are sufficient for reliable disambiguation.
    
    \item \textbf{Template B (Disambiguation-Focused):} This stricter template is triggered only when two conditions are met simultaneously:
    \begin{enumerate}
        \item The acronym is unseen (zero-shot, $a_j \in C_{\text{test} \setminus \text{train}}$); AND
        \item The candidate set $O_j$ contains ambiguous options with high morphological overlap.
    \end{enumerate}
    We quantify this overlap using a \textbf{stemming-based Jaccard similarity} threshold. When these conditions hold, the model is prone to confusion between semantically adjacent terms. Template B counteracts this by enforcing explicit reasoning steps and favoring technically specific expansions over generic ones.
\end{itemize}

\subsection{Contextual Selection}
\label{sec:selection}
The Contextual Selection module operates conditionally, activating only when the target acronym has been observed in the training set ($a_j \in C_{\text{test} \cap \text{train}}$). This corresponds to the standard few-shot path (Template A). For each such instance, we use a BM25 index \citep{robertson2009probabilistic} constructed over the training corpus to retrieve a candidate pool of contextually similar occurrences.

To ensure the prompt remains informative and unbiased, the selector applies two rigorous refinement steps:
\begin{enumerate}
    \item \textbf{Balanced Sampling:} A strategy designed to retrieve an equal distribution of positive (matching the target sense) and negative (contrastive) instances, thereby preventing the model from defaulting to the majority class.
    \item \textbf{Diversity-Aware Deduplication:} A filtering step that eliminates redundant examples using normalized text similarity, ensuring that the limited context window is utilized for semantically distinct demonstrations.
\end{enumerate}

For a given input $(a_j, t_j)$, the module returns a curated set of up to six examples. Conversely, for unseen acronyms ($a_j \in C_{\text{test} \setminus \text{train}}$), the selection phase is bypassed, and the system naturally reverts to zero-shot inference. Algorithm \ref{alg:prompt_selector} details the complete decision logic governing this selection and prompt construction process.

\begin{algorithm}[h]
\caption{Prompt Selection and Few-Shot Retrieval Strategy}
\label{alg:prompt_selector}
\begin{small}
\begin{enumerate}
    \item \textbf{Input:} Test instance $e^{\text{test}}_j = (a_j, t_j, O_j)$, threshold $\tau$, limit $N_{\text{fs}}$.
    \item \textbf{Output:} Template $T$ and few-shot set $\mathcal{F}_j$.
    \item[] 
    
    \item \textbf{Case 1: Seen acronym ($a_j \in C_{\text{test} \cap \text{train}}$)}
    \begin{itemize}
        \item Set $T \leftarrow \textsc{Template A}$.
        \item Retrieve examples via BM25 index of $E_{\text{train}}$.
        \item Apply \textit{Balanced Sampling} and \textit{Diversity Deduplication}.
        \item Set $\mathcal{F}_j \leftarrow$ top $N_{\text{fs}}$ refined examples.
        \item \textbf{Return} $(T, \mathcal{F}_j)$.
    \end{itemize}
    \item[] 
    
    \item \textbf{Case 2: Unseen acronym ($a_j \in C_{\text{test} \setminus \text{train}}$)}
    \begin{itemize}
        \item Calculate pairwise option similarity: $S_{max} = \max \mathrm{Sim}(O_{j_p}, O_{j_q})$.
        
        \item \textbf{If} $S_{max} \ge \tau$:
        \begin{itemize}
            \item Set $T \leftarrow \textsc{Template B}$ (Strict Disambiguation).
        \end{itemize}

        \item \textbf{Else:}
        \begin{itemize}
            \item Set $T \leftarrow \textsc{Template A}$ (Default).
        \end{itemize}
        \item Set $\mathcal{F}_j \leftarrow \varnothing$ (Zero-shot).
        \item \textbf{Return} $(T, \mathcal{F}_j)$.
    \end{itemize}

\end{enumerate}
\end{small}
\end{algorithm}

\subsection{Retrieval Augmented Generation (Knowledge Grounding)}
\label{sec:retrieval}
To reduce the semantic mismatch between general-purpose language models and the highly specialized vocabulary of the railway domain, we developed a dedicated Knowledge Base (KB) tailored to acronym disambiguation. This KB was constructed by aggregating and curating information from three complementary sources: publicly available railway glossaries, open-source SNCF technical documentation, and authoritative long-form expansions extracted directly from the training set. The resulting resource is structured as a dictionary in which acronyms serve as keys and their corresponding lists of validated expansions serve as values; these expansions are then injected into the prompt (see Appendix A). Together, these sources and their organization provide a reliable and domain-specific lexical foundation.

To exploit this resource effectively, we implemented a hybrid retrieval mechanism. For each acronym instance in the test set, the system queries the KB to obtain not only the candidate long forms but also their verified definitions and representative usage examples. This retrieval step acts as a grounding layer that constrains the language model’s reasoning process. By conditioning the model on explicit, domain-authenticated knowledge rather than relying solely on its parametric memory, we ensure that subsequent predictions remain aligned with the technical semantics and operational constraints of the railway context.

\subsection{Ensemble Aggregation}
\label{sec:ensemble}
The final component of the DACE framework is the Ensemble Aggregation module. To mitigate the stochastic nature of individual LLMs and reduce hallucination risks, we employ an ensemble learning strategy. This approach employs a diverse pool of $M$ candidate models, systematically filtering and combining their outputs to ensure robust decision-making.

Once the prompt template $T$ and (if applicable) the few-shot set $\mathcal{F}_j$ are finalized, the system instantiates the specific prompt $P(e^{\text{test}}_j)$ for the test instance $e^{\text{test}}_j = (a_j, t_j, O_j)$. This prompt is broadcast to the set of selected models. Formally, each LLM $\mathcal{L}_m$ (where $m \in \{1,\ldots,M\}$) maps the input to a prediction set:
\[
\hat{Y}^{(m)}_j = \mathcal{L}_m\big(P(e^{\text{test}}_j)\big),
\]
where $\hat{Y}^{(m)}_j \subseteq \{1,\ldots,|O_j|\}$ denotes the indices of the predicted options. If a model fails to identify a valid option or predicts "none," $\hat{Y}^{(m)}_j = \varnothing$. The inference stage thus yields a collection of predictions $\mathcal{\hat{Y}}_j = \{\hat{Y}^{(1)}_j, \ldots, \hat{Y}^{(M)}_j\}$.

Rather than a naive majority vote across all $M$ models, we construct the final decision $\hat{y}_j$ using a cascaded logic that prioritizes high-performing, complementary models. This process is governed by three principles:

\paragraph{1. Subset Selection:} 
We strictly limit the voting pool to a subset $\mathcal{S} \subseteq \{1,\ldots,M\}$ of models that demonstrate strong individual performance and error diversity on the validation set.

\paragraph{2. Majority Voting with Tie-Breaking:} 
For a given instance, we compute the majority consensus over $\mathcal{S}$.
\begin{itemize}
    \item \textbf{Standard Case:} If $|\mathcal{S}|$ is odd, or if $|\mathcal{S}|$ is even but no tie occurs, the final output is the majority prediction:
    \[
    \hat{y}_j = \mathrm{MajVote}\big(\{\hat{Y}^{(m)}_j : m \in \mathcal{S}\}\big).
    \]
    \item \textbf{Tie-Breaking:} If $|\mathcal{S}|$ is even and a perfect tie occurs (e.g., a 2-vs-2 split), we query a designated \textit{tie-breaker model} $\mathcal{L}_{m^\star}$ (where $m^\star \notin \mathcal{S}$). This model is selected specifically for its complementary reasoning patterns relative to $\mathcal{S}$. The vote is then recomputed over the extended set $\mathcal{S} \cup \{m^\star\}$ to force a decision.
\end{itemize}

\paragraph{3. Competence Fallback:} 
In rare cases of \textit{high divergence}, where the ensemble fails to reach a significant consensus (i.e., multiple conflicting predictions with low support), the system discards the vote and falls back to the single model with the highest historical accuracy, denoted as $m_{\text{best}}$:
\[
\hat{y}_j = \hat{Y}^{(m_{\text{best}})}_j.
\]

\section{Experiments}

\subsection{Dataset}
\label{sec:dataset}
The TextMine'26 challenge dataset comprises two partitions: a training set of 492 annotated examples and a test set of 519 examples. Each example consists of a short segment of railway technical text containing a single target acronym, for which the task is to predict the correct expanded form from a provided set of candidate options.

\begin{figure}[t]
\begin{center}
 \includegraphics[width=10cm]{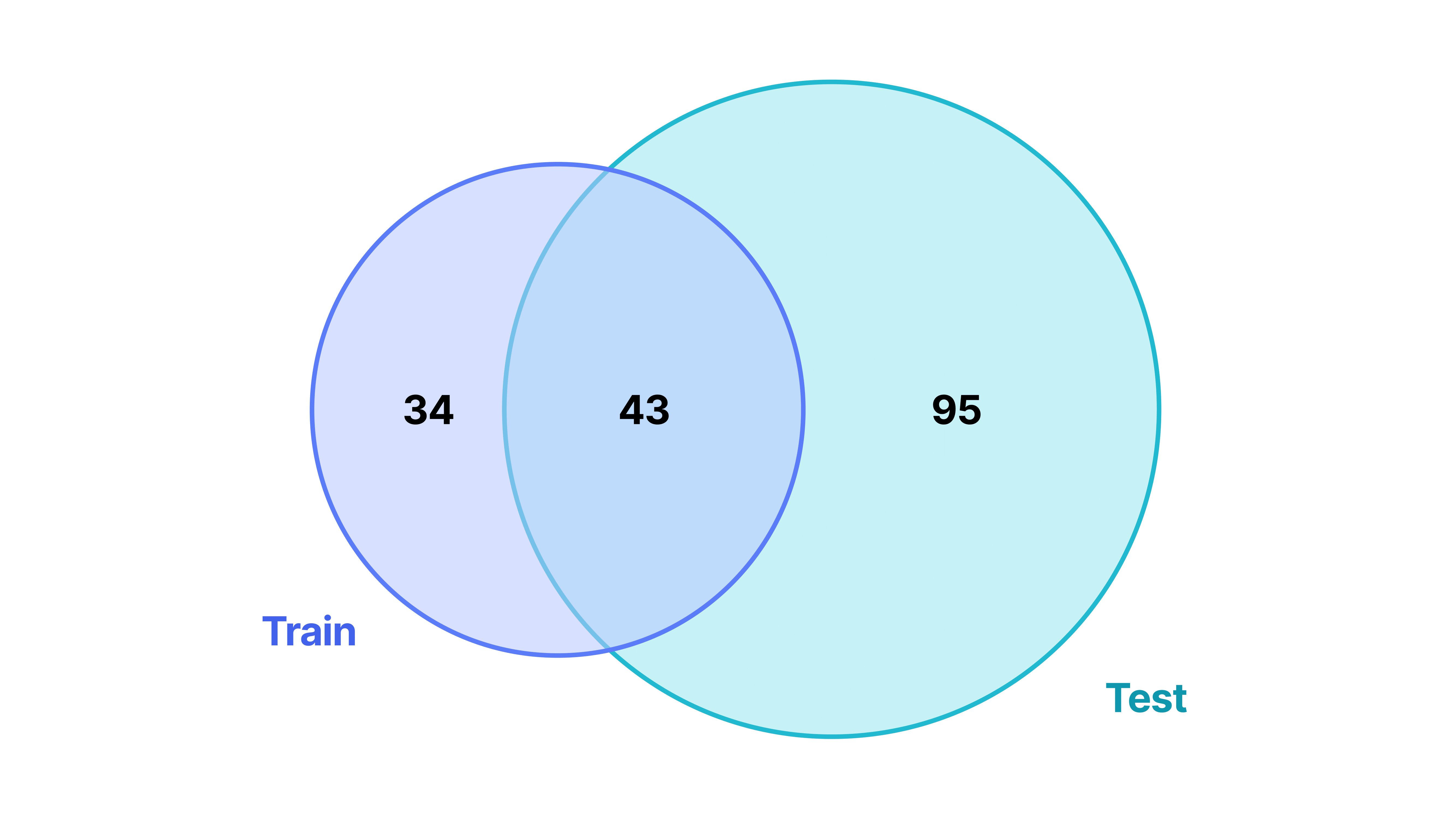}
 \caption{Distribution of unique acronyms across the training and test sets.} \label{fig_unique_acronyms_chart}
\end{center}
\end{figure}

The dataset contains a diverse collection of acronyms with varying levels of ambiguity. Figure \ref{fig_unique_acronyms_chart} shows the number of unique acronyms in each split. This distribution reveals a critical challenge: while certain acronyms appear in both partitions, a substantial proportion of test acronyms (95 unique acronyms) are absent from the training data.

\begin{figure}[t]
    \centering
    \begin{subfigure}[b]{0.56\textwidth}
        \centering
        \includegraphics[width=1.0\textwidth]{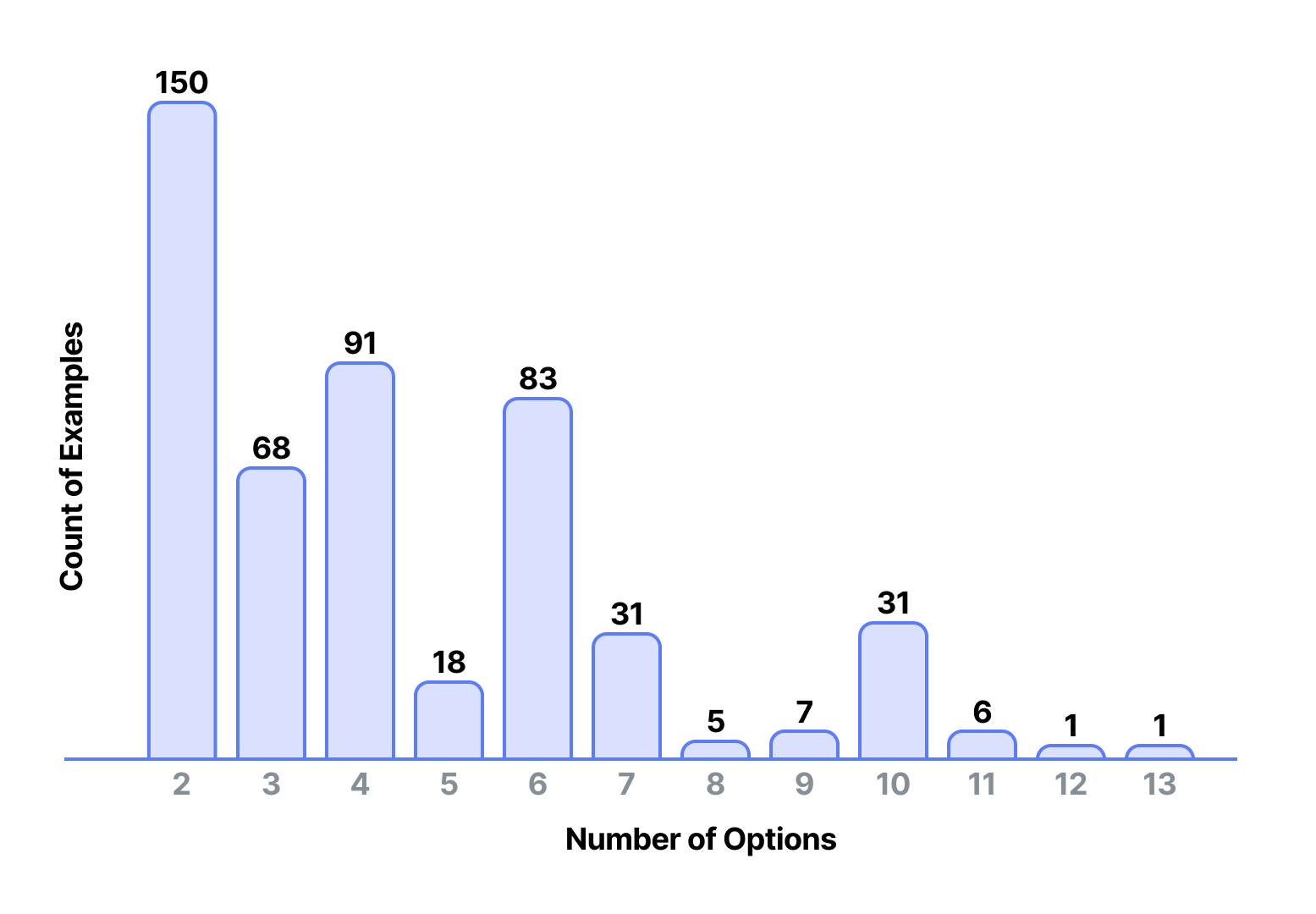}
        \caption{Distribution of the number of candidate expansions per training example.}
        \label{fig_options_chart}
    \end{subfigure}
    \hfill
    \begin{subfigure}[b]{0.40\textwidth}
        \centering
        \includegraphics[width=1.0\textwidth]{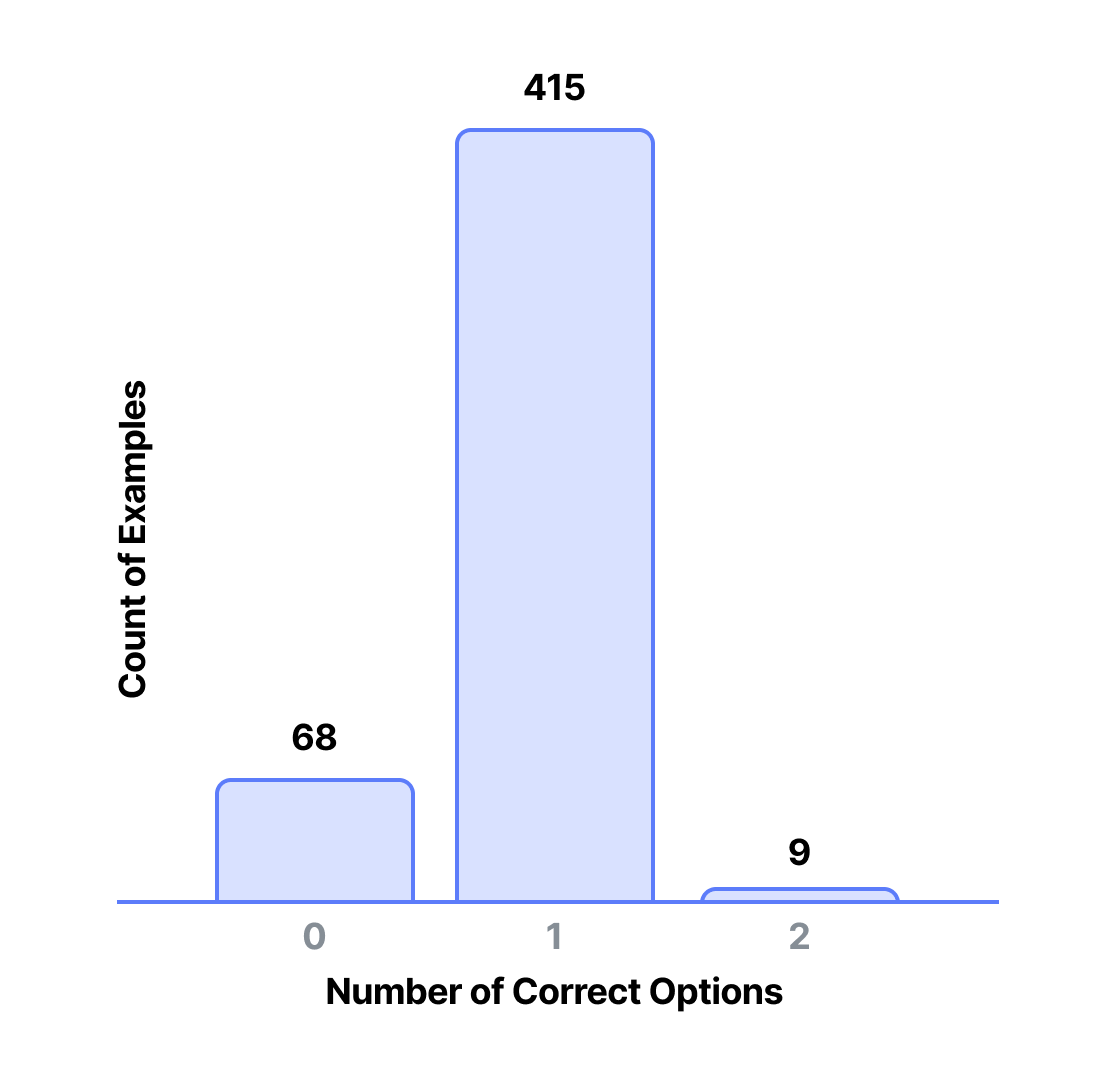}
        \caption{Distribution of the number of correct candidate expansions.}
        \label{fig_correct_options_chart}
    \end{subfigure}
    \caption{Analysis of candidate and correct expansion distributions in the training set. Figure \ref{fig_options_chart} shows the variability in the number of candidate options, and Figure \ref{fig_correct_options_chart} shows the distribution of correct answers.}
    \label{fig_ambiguity_analysis}
\end{figure}

The number of candidate expansions per example varies considerably, ranging from 2 to 13 options (Figure \ref{fig_options_chart}). This wide range reflects heterogeneous levels of ambiguity across acronyms and introduces substantial variation in task difficulty: some instances present binary choices, while others require selection from large candidate pools. This variability demands flexible disambiguation strategies capable of handling both extremes without performance degradation.

Each example may contain zero, one, or two correct expansions, reflecting real-world scenarios where an acronym may lack a valid expansion among the proposed candidates or may have multiple acceptable long forms depending on context. Figure \ref{fig_correct_options_chart} presents the distribution of correct answers across the dataset. Notably, 68 training examples (13.8\%) contain no correct option among the provided candidates, modeling realistic situations where the true expansion is absent from the candidate list. Additionally, 9 examples exhibit multiple correct expansions, requiring systems capable of multi-label output or context-dependent sense detection. The presence of zero-correct examples increases task complexity, as models must avoid overconfidence and accurately identify when no candidate matches the contextual usage.

The dataset exhibits several characteristics that fundamentally inform methodological design decisions. First, it features numerous low-resource and long-tail acronyms, with many appearing infrequently in training data or exclusively at test time. The presence of 95 test-only acronyms indicates that approaches relying heavily on per-acronym supervised learning face significant generalization challenges. Second, the variable candidate set sizes necessitate prompting and selection strategies that maintain robust performance across both small and large candidate pools. Third, the task requires models to handle edge cases where correct answers may be absent entirely or where multiple valid expansions coexist. Finally, the domain-specific nature of the data presents unique challenges: examples are drawn from prescriptive railway documentation and feature highly specialized, context-dependent meanings that often diverge from general usage. This domain specificity motivates the incorporation of external domain knowledge through retrieval mechanisms and the use of context-aware few-shot examples to guide models toward domain-appropriate interpretations.

\subsection{Experimental Setup}
All experiments were conducted using the official APIs provided by Anthropic and Google. Based on preliminary validation on the training set, we selected three high-performing LLMs for our ensemble: \textbf{Claude 3.5 Sonnet (v2-20241022)}, \textbf{Claude 4.1 Opus (v4.1-20250805)}, and \textbf{Gemini 2.5 Pro}.

To ensure reproducibility and minimize stochastic variance, we set the temperature parameter to $\tau=0$ (greedy decoding) for all inference calls. We enforced a strict output schema using JSON mode, requiring the model to return a dictionary mapping each candidate label to a boolean verdict (e.g., \texttt{\{"A": true, "B": false\}}). This constrained generation eliminated parsing errors and ensured that the models focused solely on the binary classification of candidates.

To facilitate reproducibility and further research, our complete code, including prompt templates and environment configurations, is publicly available at: \url{https://github.com/elmontaser1998/TextMine_2026}.

\subsection{Main Results}
The comparative performance of the participating teams is summarized in Table~\ref{tab_ranking}. Our proposed method, submitted under the team name \textbf{Cursive}, achieved the top position on both the public and private leaderboards.

\begin{table}[ht]
\setlength{\tabcolsep}{12pt}
\begin{tabular}{llcc}
    \hline\hline
    \textbf{Rank} & \textbf{Team} & \textbf{Public F1} & \textbf{Private F1} \\
    \hline
    \textbf{1} & \textbf{Cursive (Ours)} & \textbf{0.9017} & \textbf{0.9069} \\
    2 & Mokipo\_ & 0.8806 & 0.8814 \\
    3 & AR & 0.8663 & 0.8632 \\
    4 & EDF R\&D & 0.8394 & 0.8604 \\
    5 & David & 0.7916 & 0.8419 \\
    6 & Mehdinho & 0.7846 & 0.8108 \\
    7 & yqnis & 0.7768 & 0.7867 \\
    8 & maksimko & 0.8129 & 0.7812 \\
    \hline
\end{tabular}
\centering
\caption{Final standings on the TextMine'26 Public and Private test sets. Our \textbf{Cursive} submission (using DACE) outperforms the runner-up by a significant margin.}
\label{tab_ranking}
\end{table}

We achieved a final Private F1 score of \textbf{0.9069}, outperforming the second-best team by approximately 2.5 percentage points. A critical observation is the stability of our model: while many competitors saw performance fluctuations between the Public and Private splits, our score remained consistent (and even slightly improved on the Private set). This indicates that the \textbf{DACE} framework prevents overfitting to specific training examples and generalizes effectively to unseen data.

\subsection{Ablation Study}
To quantify the contribution of each component in our pipeline, we conducted an ablation study (Table~\ref{tab_ablation}). We systematically compared baseline single-prompt approaches against the full DACE pipeline to isolate the impact of dynamic strategy selection, model architecture, and ensemble aggregation.

\begin{table}[ht]
\centering
\setlength{\tabcolsep}{10pt}
\begin{tabular}{lcc}
    \hline\hline
    \textbf{Method} & \textbf{Public F1} & \textbf{Private F1} \\
    \hline
    Claude 3.5 Sonnet (Template A only) & 0.8407 & 0.8571 \\
    Claude 3.5 Sonnet (Template B only) & 0.8292 & 0.8368 \\
    Claude 3.5 Sonnet + Dynamic Prompting & 0.8876 & 0.8940 \\
    Claude 4.1 Opus + Dynamic Prompting & 0.8778 & 0.8852 \\
    \textbf{DACE (Full Ensemble)} & \textbf{0.9017} & \textbf{0.9069} \\
    \hline
\end{tabular}
\caption{Ablation study demonstrating the incremental value of Dynamic Prompting and Ensembling. Baseline models use static zero-shot or few-shot prompts.}
\label{tab_ablation}
\end{table}

Results indicate that static prompting strategies are insufficient in isolation. The "Template A only" baseline, which uses standard few-shot learning, achieved a Private F1 of 0.8571. This approach struggles particularly with the 95 unseen test acronyms, where the absence of training examples often leads to frequency-biased errors. Conversely, the "Template B only" approach (Private F1 0.8368) proved too rigid. While it successfully reduces hallucinations for highly ambiguous terms by enforcing strict definition matching, it lacks the contextual nuance provided by few-shot demonstrations, causing it to underperform on standard cases where context is key.

The introduction of Dynamic Prompting provided the most significant boost, raising the Private F1 to 0.8940. This module intelligently switches between templates based on acronym visibility and candidate ambiguity. By automatically routing unseen or highly overlapping instances to the stricter Template B while reserving Template A for standard cases, the system effectively bridges the generalization gap. This adaptability is crucial for handling the dataset's significant portion of low-resource acronyms that static prompts fail to capture.

Finally, the Ensemble Aggregation module achieved the top Private F1 of 0.9069. By aggregating predictions from diverse models, specifically combining the reasoning of Claude 3.5 Sonnet and Claude 4.1 Opus, the ensemble corrects idiosyncratic hallucinations and mitigates stochastic variance. This stability is evidenced by the consistent performance between Public (0.9017) and Private (0.9069) evaluations, confirming that the framework robustly resolves genuine ambiguity rather than overfitting to specific leaderboard splits.

\section{Conclusion}
This work introduced DACE, a modular framework that leverages dynamic prompt engineering, retrieval augmentation, contextual demonstration selection, and ensemble aggregation to address the challenges of acronym disambiguation in highly specialized railway documentation. Our design choices were grounded in two observations drawn from the dataset analysis: first, the substantial presence of low-resource and unseen acronyms requires methods that generalize beyond supervised training examples; second, the heterogeneity of candidate sets and domain-specific semantics demands explicit grounding in authoritative technical knowledge. These considerations motivated a hybrid approach that integrates LLM reasoning with structured retrieval mechanisms and carefully controlled few-shot prompting.

Experimental results confirm the effectiveness of this strategy, with DACE achieving the highest F1 score in the TextMine’26 competition and maintaining stable performance between public and private evaluations. The ablation study further demonstrates that each module contributes meaningfully to robustness. Dynamic prompting reduces failure cases on ambiguous acronyms, retrieval improves factual alignment with railway terminology, and ensembling mitigates stochastic variance across models. Together, these results validate the core hypothesis underlying our methodology, namely that adaptive prompt construction coupled with external knowledge grounding can compensate for limited supervision and strong domain shift.

Beyond the competition setting, the generality of the framework suggests broader applicability. Its components are not tied to a specific language model, domain, or acronym inventory, which opens opportunities for transfer to other technical sectors where terminology is dense, polysemous, and weakly annotated. Future research will focus on extending DACE to multilingual scenarios, integrating structured knowledge graphs as an additional retrieval source, and evaluating the system on larger open benchmarks such as GLADIS. A systematic exploration of automatic prompt optimization and reinforcement-driven example selection also constitutes a promising direction. Overall, our results illustrate the practical value of combining LLM adaptability with principled information retrieval techniques for domain-specific disambiguation tasks.

\section*{Appendix A: Prompt Templates Examples}

\subsection*{Template A}

\begin{Verbatim}[breaklines=true]
You are an expert in acronym disambiguation, specializing in French railway (SNCF) contexts. Always prioritize domain-specific knowledge and the provided tie-breakers.

Read the following text carefully:
EP JT42CWRM-120 Tout le RT / compatibilité :OUI/restriction : néant] D/2011/003439/

Determine what the acronym "EP" stands for **in this specific context**. Select zero or more options, but only if they precisely match the context—do not guess or over-select. If no option fits perfectly, select none.

Choose from the options below (zero or more may be correct).
Options:
A. Enquête publique
B. Exercice pratique
C. Embranchement particulier
D. canalisation des Eaux Pluviales
E. Entretien Professionnel :   Il constitue un moment clé d'échange, consacré aux perspectives d'évolution professionnelle, notamment en termes de qualification et d'emploi. Il permet de faire le point sur les acquis des formations déjà suivies et d'identifier les besoins de développement des compétences du salarié. C'est une obligation légale. Il est préconisé de le réaliser au moment de l'entretien individuel d'appréciation (EIA) ou du rendez-vous professionnel individuel annuel (RPIA).
F. Enclenchement de proximité
G. Empreinte de patinage
H. Équipe-projet
I. Etablissement public
J. Etude Préliminaire
K. Échange Prenant


Valid definitions (for reference — ignore any that do not appear in Options):
- l'Entretien Professionnel a remplacé depuis 2014 l'EIF (Entretien individuel formation). Il est idéalement mené à la suite de l'EIA afin :_x000D_
- d'identifier les besoins de formation des agents afin de favoriser leur évolution professionnelle._x000D_
- de faire un point sur les acquis des formations suivies._x000D_
- de développer les compétences des agents en leur proposant une formation adaptée.
- Embranchement particulier
- Empreinte de patinage
- Exercice pratique
- Eaux pluviales
- Etablissement public
- Enquête publique
- Enclenchement de proximité
- Équipe-projet

Examples (balanced & similar):
{"text": "EIC établissement infrastructure circulation du service chargé de la gestion du trafic et des circulations sur le RFN EP embranchement particulier EPSF établissement public de sécurité ferroviaire GI gestionnaire d'infrastructure  ", "acronym": "EP", "options": {"Enquête publique": false, "Entretien Professionnel :   Il constitue un moment clé d'échange, consacré aux perspectives d'évolution professionnelle, notamment en termes de qualification et d'emploi. Il permet de faire le point sur les acquis des formations déjà suivies et d'identifier les besoins de développement des compétences du salarié. C'est une obligation légale. Il est préconisé de le réaliser au moment de l'entretien individuel d'appréciation (EIA) ou du rendez-vous professionnel individuel annuel (RPIA).": false, "Etude Préliminaire": false, "canalisation des Eaux Pluviales": false, "Exercice pratique": false, "Enclenchement de proximité": false, "Équipe-projet": false, "Embranchement particulier": true, "Échange Prenant": false, "Etablissement public": false, "Empreinte de patinage": false}}

{"text": "AC attestation de compatibilité COGC centre opérationnel de gestion des circulations DT double traction EM engin moteur EP embranchement particulier ", "acronym": "EP", "options": {"Embranchement particulier": true, "Équipe-projet": false, "Enclenchement de proximité": false, "Enquête publique": false, "Empreinte de patinage": false, "Échange Prenant": false, "Etude Préliminaire": false, "Exercice pratique": false, "Etablissement public": false, "canalisation des Eaux Pluviales": false, "Entretien Professionnel :   Il constitue un moment clé d'échange, consacré aux perspectives d'évolution professionnelle, notamment en termes de qualification et d'emploi. Il permet de faire le point sur les acquis des formations déjà suivies et d'identifier les besoins de développement des compétences du salarié. C'est une obligation légale. Il est préconisé de le réaliser au moment de l'entretien individuel d'appréciation (EIA) ou du rendez-vous professionnel individuel annuel (RPIA).": false}}

{"text": "CRUAS PL ……..………………………..EP de l’E.D.F. ……………………...…….EP des Ciments Lafarge………. …………Bif. Raccordement Sud de LA VOULTE..Etablissements", "acronym": "EP", "options": {"canalisation des Eaux Pluviales": false, "Etablissement public": false, "Empreinte de patinage": false, "Équipe-projet": false, "Entretien Professionnel :   Il constitue un moment clé d'échange, consacré aux perspectives d'évolution professionnelle, notamment en termes de qualification et d'emploi. Il permet de faire le point sur les acquis des formations déjà suivies et d'identifier les besoins de développement des compétences du salarié. C'est une obligation légale. Il est préconisé de le réaliser au moment de l'entretien individuel d'appréciation (EIA) ou du rendez-vous professionnel individuel annuel (RPIA).": false, "Enquête publique": false, "Etude Préliminaire": false, "Embranchement particulier": true, "Échange Prenant": false, "Enclenchement de proximité": false, "Exercice pratique": false}}

{"text": "AUTOR autorail CLE consigne locale d'exploitation COGC centre opérationnel de gestion des circulations EP embranchement particulier EPSF établissement public de sécurité ferroviaire ", "acronym": "EP", "options": {"Exercice pratique": false, "Enquête publique": false, "Équipe-projet": false, "Empreinte de patinage": false, "Etablissement public": false, "canalisation des Eaux Pluviales": false, "Etude Préliminaire": false, "Enclenchement de proximité": false, "Échange Prenant": false, "Embranchement particulier": false, "Entretien Professionnel :   Il constitue un moment clé d'échange, consacré aux perspectives d'évolution professionnelle, notamment en termes de qualification et d'emploi. Il permet de faire le point sur les acquis des formations déjà suivies et d'identifier les besoins de développement des compétences du salarié. C'est une obligation légale. Il est préconisé de le réaliser au moment de l'entretien individuel d'appréciation (EIA) ou du rendez-vous professionnel individuel annuel (RPIA).": false}}

{"text": "EP embranchement particulier EPSF établissement public de sécurité ferroviaire GSM-GFU global system for mobil - groupe fermé d'utilisateurs IPCS installation permanente de contre sens KVB contrôle de vitesse par balises ", "acronym": "EP", "options": {"Équipe-projet": false, "Exercice pratique": false, "Embranchement particulier": true, "Etablissement public": false, "Empreinte de patinage": false, "Enclenchement de proximité": false, "canalisation des Eaux Pluviales": false, "Échange Prenant": false, "Enquête publique": false, "Entretien Professionnel :   Il constitue un moment clé d'échange, consacré aux perspectives d'évolution professionnelle, notamment en termes de qualification et d'emploi. Il permet de faire le point sur les acquis des formations déjà suivies et d'identifier les besoins de développement des compétences du salarié. C'est une obligation légale. Il est préconisé de le réaliser au moment de l'entretien individuel d'appréciation (EIA) ou du rendez-vous professionnel individuel annuel (RPIA).": false, "Etude Préliminaire": false}}

{"text": "A301.3 Dispositions communes aux deux sens  Ne pas dépasser  30 km/h au franchissement des ponts situés aux km :  - 269,458 entre EP La Vauvelle PL et Epiry-Montreuillon PL ", "acronym": "EP", "options": {"Équipe-projet": false, "Etude Préliminaire": false, "Empreinte de patinage": false, "Enclenchement de proximité": false, "Etablissement public": false, "Échange Prenant": false, "Enquête publique": false, "Embranchement particulier": true, "Entretien Professionnel :   Il constitue un moment clé d'échange, consacré aux perspectives d'évolution professionnelle, notamment en termes de qualification et d'emploi. Il permet de faire le point sur les acquis des formations déjà suivies et d'identifier les besoins de développement des compétences du salarié. C'est une obligation légale. Il est préconisé de le réaliser au moment de l'entretien individuel d'appréciation (EIA) ou du rendez-vous professionnel individuel annuel (RPIA).": false, "canalisation des Eaux Pluviales": false, "Exercice pratique": false}}

Think step by step **internally**, considering the text, options, few-shots, and tie-breakers.
Output ONLY a JSON object with booleans for each label, e.g., {"A": true, "B": false}.
\end{Verbatim}

\subsection*{Template B}

\begin{Verbatim}[breaklines=true]
You are an expert in acronym disambiguation in French railway (SNCF) contexts.

Task:
Determine which option(s) correctly expand and describe the acronym in the given text.

Selection Rules:
1. Use the surrounding text to understand the functional and technical context.
2. If multiple options correspond to the SAME underlying meaning/expansion:
   - Select ONLY ONE of them.
   - Prefer the option that is more precise, more technical, or more complete
     (usually the longer, fully formulated description).
3. If one option is a short label and another is the same concept with a detailed explanation,
   choose ONLY the detailed one.
4. If none of the options truly match the context, select none.
5. Do NOT invent new meanings.

Text:
atteignant 10 minutes. _x0001_ Ligne autorisée à la charge D pour les trains MA et ME 120. § 1.2 : Dispositions particulières Metz-Ville - Stiring-Wendel (Sarrebruck) : _x0001_ Voies banalisées : 

Acronym: MA

Options:
A. Marche arrêt
B. movement authority
C. Marchandise
D. Maladie
E. Maintenance et appui
F. Manuel / automatique


Valid definitions (for reference — ignore any that do not appear in Options):
- movement authority
- Maladie
- Maintenance et appui
- Marche arrêt
- Manuel / automatique

Output ONLY a JSON object with booleans for each label (A, B, C, ...).
A label is true if and only if its option is selected according to the rules above.
\end{Verbatim}

\bibliographystyle{rnti}
\bibliography{biblio_exemple}

\end{document}